\documentclass[runningheads]{llncs}
\usepackage[T1]{fontenc}
\usepackage{graphicx}
\usepackage{multirow}
\usepackage{amsmath}
\usepackage{amsfonts}
\usepackage{xcolor}
\usepackage{booktabs}
\usepackage{etoolbox}
\usepackage{authblk}
\usepackage{array}
\usepackage{wrapfig}
\usepackage{fontawesome} 
\definecolor{mygreen}{RGB}{114, 240, 126}
\newcommand{\equalcontrib}{\textsuperscript{*}}
\newcommand{\Envelope}{\textsuperscript{\faEnvelopeO}}
\begin{document}
\title{Structure-aware World Model for Probe Guidance via Large-scale Self-supervised Pre-train}
\titlerunning{Structure-aware World Model for Probe Guidance}

\author{
    Haojun Jiang\inst{1,2}\equalcontrib \and
    Meng Li\inst{2}\equalcontrib\and
    Zhenguo Sun\inst{2}\equalcontrib \and
    Ning Jia\inst{2}\and
    Yu Sun\inst{2}\and
    Shaqi Luo\inst{2}\and
    Shiji Song\inst{1}\and
    Gao Huang\inst{1,2}\Envelope
}

\authorrunning{H. Jiang et al.}

\institute{Department of Automation, BNRist, Tsinghua University, Beijing, China\\
\and
Beijing Academy of Artificial Intelligence, Beijing, China\\
\email{jianghaojunthu@163.com}, \email{s10133679@163.com}, \email{hitsunzhenguo@gmail.com}, \email{gaohuang@tsinghua.edu.cn}
}
\maketitle              
\renewcommand{\thefootnote}{}
\footnotetext[1]{\equalcontrib These authors contributed equally to this work. This work was done while Haojun Jiang was an intern at Beijing Academy of Artificial Intelligence.}
\footnotetext[2]{\Envelope Corresponding author.}

\begin{abstract}
The complex structure of the heart leads to significant challenges in echocardiography, especially in acquisition cardiac ultrasound images.
Successful echocardiography requires a thorough understanding of the structures on the two-dimensional plane and the spatial relationships between planes in three-dimensional space.
In this paper, we innovatively propose a large-scale self-supervised pre-training method to acquire a cardiac structure-aware world model.
The core innovation lies in constructing a self-supervised task that requires structural inference by predicting masked structures on a 2D plane and imagining another plane based on pose transformation in 3D space.
To support large-scale pre-training, we collected over 1.36 million echocardiograms from ten standard views, along with their 3D spatial poses.
In the downstream probe guidance task, we demonstrate that our pre-trained model consistently reduces guidance errors across the ten most common standard views on the test set with 0.29 million samples from 74 routine clinical scans, indicating that structure-aware pre-training benefits the scanning.
\end{abstract}
\section{Introduction}
\begin{figure}[!tp]%
\centering
\includegraphics[width=\textwidth]{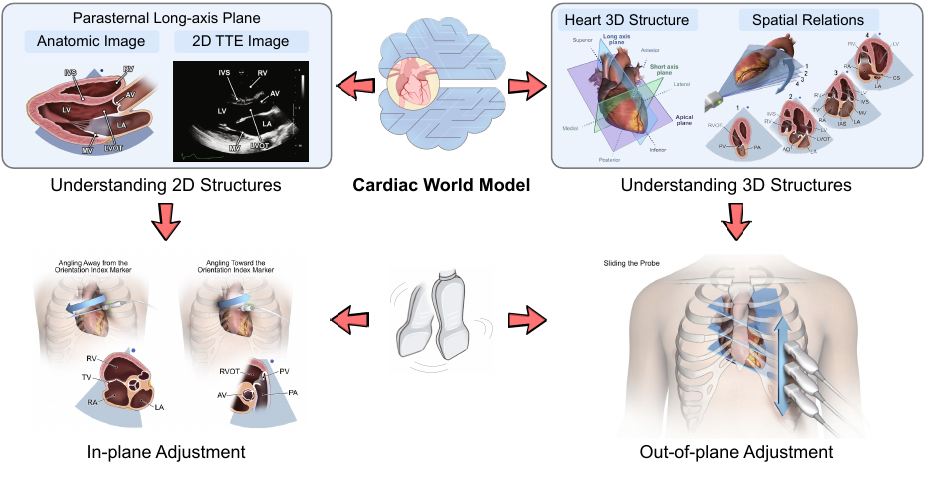}
\vspace{-6mm}
\caption{
\textbf{Diagram illustrating the capabilities of a cardiac world model.}
We aim to develop a cardiac world model that can understand both two-dimensional and three-dimensional structures.
\textbf{(Left)} The world model needs to recognize various structures in two-dimensional planes and understand their spatial relationships for in-plane probe adjustment.
\textbf{(Right)} Understanding the three-dimensional structure of the heart, specifically the spatial relationships between different planes, is crucial for out-of-plane probe adjustment.
The images used in the diagram are sourced from \cite{mitchell2019guidelines}.
}
\label{fig1}
\end{figure}

Cardiovascular diseases are the leading cause of death in worldwide~\cite{roth2017global,song2020global}. 
Echocardiography is the most commonly used method in clinical practice to assess heart conditions. 
However, the structure of the heart is extremely complex. 
According to \cite{mitchell2019guidelines}, up to seven anatomical structures need to be identified in a single plane, such as the Parasternal Long-axis Plane (Fig.~\ref{fig1} Left). 
Recognizing these anatomical structures is crucial for diagnosis, and understanding their spatial relationships in a two-dimensional plane also helps the sonographer fine-tune the probe to obtain the best quality images.
Additionally, up to 27 different planes need to be examined, requiring the sonographer to understand the spatial relationships between these planes in three-dimensional space and to finely adjust the ultrasound probe's position to reach the target location (Fig.~\ref{fig1} Right).
Due to the aforementioned reasons, cardiac ultrasound examinations are extremely challenging. 
This also results in a long training period for ultrasound medical personnel, as they need to spend a lot of time familiarizing themselves with both 2D and 3D structures. 
Consequently, there is a significant talent shortage in this field, especially in regions with scarce medical resources, such as Africa.

With technological advancements~\cite{dosovitskiy2020image,he2016deep,huang2017densely,wang2021adaptive,yan2023multi,yang2021condensenet}, AI algorithms has shown great potential in improving the efficiency of echocardiography.
For example, Ouyang et al.~\cite{ouyang2020video} proposed a video-based deep learning algorithm, \textit{i.e.,} EchoNet-Dynamic, that can automatically make accurate assessments of cardiac function.
More importantly, with the rise of large language models~\cite{brown2020language,radford2018improving,radford2019language,thirunavukarasu2023large} and multi-modal learning~\cite{jiang2022pseudo,jiang2022cross,li2021align,radford2021learning}, the interpretation of echocardiograms~\cite{christensen2024vision} has shown increasing improvement, demonstrating excellent performance in tasks such as pulmonary artery pressure estimation, left ventricular hypertrophy, heart failure, and left atrial enlargement.
These AI-assisted diagnostic tools have demonstrated performance almost comparable to human experts. 
However, this comes with a prerequisite: the acquisition of high-quality echocardiograms.
In regions with scarce medical resources, there are often no sonographers available to obtain high-quality echocardiograms. 
In such cases, the powerful capabilities of these AI-assisted diagnostic tools cannot be fully utilized.
Few works~\cite{droste2020automatic,jiang2024cardiac,narang2021utility} have focused on how to use AI technology to assist inexperienced sonographers in accurately acquiring target planes.
Recently, Jiang et al.~\cite{jiang2024cardiac} proposed a cardiac dreamer, which only focuses on supervised learning of the 3D-structure of the heart and serves as a "heart map" for the probe guidance task.
This work demonstrates a potential AI-assisted scanning method that is expected to improve the scanning skills of novices.

According to the clinical experience of sonographers, understanding both the 2D and 3D structures is crucial for efficient scanning.
For example, when you need to adjust the probe's viewing angle on a particular plane to capture specific anatomical structures, you must have a good understanding of the spatial positions of those anatomical structures on that plane (Fig.\ref{fig1} Left).
Undoubtedly, when navigating between different planes, it is essential for sonographers to grasp the spatial relationships within the three-dimensional space (Fig.\ref{fig1} Right).
Thus, in this paper, we propose a 2D-3D joint structure-aware pre-training framework to obtain a data-driven cardiac world model that benefits ultrasound scanning.
Specifically, we require the world model to learn important spatial relationships in the following ways:
(1) for understanding two-dimensional structures, we use a masking approach that requires the world model to predict features at adjacent positions in the two-dimensional plane;
(2) for understanding three-dimensional structures, we provide information on the positional changes of two planes in 3D space, requiring the world model to predict the features of the target plane after the positional change.
We further collected expert operational data on acquiring the most common ten standard planes from 364 routine clinical scans, resulting in 1.36 million sample pairs gathered by three certified sonographers, to enhance the model's learning of generalizable cardiac structure knowledge.
Results on the downstream probe guidance task indicate that the proposed pre-training method learns useful knowledge for assisting in the acquisition of echocardiograms.
\section{Method}
\begin{figure}[!tp]%
\centering
\includegraphics[width=\textwidth]{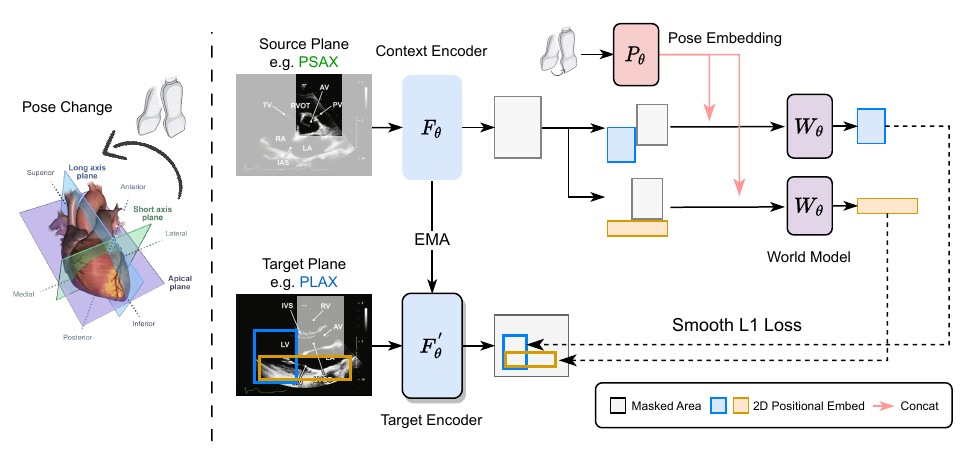}
\caption{
\textbf{Diagram illustrating the pre-training method and downstream task.}
The world model and encoder are required to predict features on the target plane based on the spatial relationships in both two-dimensional and three-dimensional spaces.
}
\label{fig2}
\end{figure}

In this section, we describe the proposed structure-aware pre-training framework, illustrated in Fig.~\ref{fig2}.
We first introduce the prior I-JEPA work~\cite{assran2023self} , on which we based our method, in Section \ref{method:pre}. 
Next,  we discuss how we construct a pre-training framework that simultaneously learns 2D and 3D structural information by introducing three-dimensional spatial information in Section \ref{method:structure}.

\subsection{Preliminary}
\label{method:pre}
In the context of echocardiogram analysis, accurately understanding 2D structural information is fundamental for making correct diagnoses and conducting efficient scans.
Recently, Assran et al. proposed a Joint-Embedding Predictive Architecture (I-JEPA)~\cite{assran2023self} to learn highly semantic image representation. 
The key idea for I-JEPA is learning the representation of images by predicting
features of target blocks based on the non-overlapping context block and the positional embedding in the same image.
This paradigm requires the model to understand the spatial relationships of different semantic structures in the 2D plane. 
For example, if a context block contains the head of a dog, it is likely that the body of the dog will be located below the context block.
For echocardiogram analysis, this paradigm also enables the model to learn the spatial relationships of fine structures in the two-dimensional plane.
For instance, the left ventricle (LV) is located below the interventricular septum (IVS), as shown on the left side of Fig.\ref{fig1}.
Therefore, it is highly suitable for modeling two-dimensional structural information in cardiac ultrasound images.
Next, we briefly introduce the modeling and training method.

\textbf{Targets.} Specifically, the I-JEPA model employs a Vision Transformer~\cite{dosovitskiy2020image}.
Firstly, the input image $\mathbf{I} \in \mathbb{R}^{H \times W}$ is divided into $N$ non-overlapping patches, which are further processed by the target encoder $F^{'}_{\theta}$.
Then, we select a portion of the patches to form $M$ target blocks, which may have overlapping regions.
We denote the ground-truth target blocks' features as $\mathbf{Y}^{t}_{i} \in \mathbb{R}^{L_t \times C}, i \in \{1,2,\cdots,M\}$.

\textbf{Context and Prediction.} For the context features, patches not belonging to the target blocks are randomly selected and input into the context encoder to obtain features $\mathbf{Z}_c \in \mathbb{R}^{L_c \times C}$.
Subsequently, the predictor $W_\theta$ (world model) generates the features of target blocks based on the context features and positional embeddings which indicate the location of context and target blocks.
We denote the predicted target blocks' feature as $\hat{\mathbf{Y}}^{t}_{i} \in \mathbb{R}^{L_t \times C}, i \in \{1,2,\cdots,M\}$.

\textbf{Loss.} With the ground-truth and predicted features of target blocks, the model is optimized using the following loss:
\begin{align}
    \mathcal{L}_{total}&= \frac{1}{M} \sum_{i=1}^{M} \sum_{j=1}^{L_t} \mathcal{L}_{\mathrm{SmoothL1}}(\mathbf{Y}^{t}_{i,j}, \hat{\mathbf{Y}}^{t}_{i,j}).
\label{eq:loss}
\end{align}

\subsection{2D-3D Joint Structure-aware Pre-training}
\label{method:structure}
Echocardiography involves acquiring high-quality echocardiogram and subsequently conducting analysis and diagnosis based on these images.
However, previous researches~\cite{christensen2024vision,ouyang2020video} have primarily focused on understanding and analyzing the two-dimensional plane, neglecting how AI can assit in acquiring high-quality echocardiogram.
While understanding the 2D structure of a plane can aid in scanning, especially when some features of the standard view are already visible, this alone is insufficient. 
Particularly when transitioning from one standard view to another, a deep understanding of the heart's 3D structure is essential.
Therefore, we propose a 2D-3D joint structure-aware pre-training framework, as illustrated in Fig.\ref{fig2}.
The core insight is predicting the visual features of structures at target locations based on given 2D and 3D positional conditions, thereby learning the mapping relationship between spatial positions and visual features.
Next, we provide details of the modeling and training method.

\textbf{Input.}
Given a source image $\mathbf{I}^{s} \in \mathbb{R}^{H \times W}$, we select a target image $\mathbf{I}^{t} \in \mathbb{R}^{H \times W}$ from the same individual's images.
To ensure sufficient spatial pose variation between the two images, there must be an interval of at least 150 frames between their sequence numbers.
Then, both images are divided into $N$ non-overlapping patches $\mathbf{I}^{s}_{p}, \mathbf{I}^{t}_{p} \in \mathbb{R}^{N \times h \times w}$.

\textbf{Context and Target.}
First, we randomly select a rectangular region from the patches of the source image $\mathbf{I}^{s}$ as the context information. 
These patches are then fed through the context encoder $F_\theta$, resulting in the context feature $\mathbf{Z}^s$:
\begin{align}
    \mathbf{Z}^s &= F_\theta(\mathbf{I}^{s}_{p} \odot \mathbf{B}^{s}), \ \ \mathbf{Z}^s \in \mathbb{R}^{L_s \times C},
\end{align}
where $\mathbf{B}^{s} \in \mathbb{R}^{N \times 1 \times 1}$ is a binary mask indicating the selected patches, $L_s$ is the number of context patches, and $C$ is the hidden dimension.
Next, we select $M$ non-overlapping regions on the target image as the target blocks, requiring the model to understand both the three-dimensional structure of the heart and the spatial relationships of heart structures in the two-dimensional plane.
If we choose regions on the target image that have the same spatial position as the context block, the model only needs to understand the three-dimensional spatial relationships.
The target features are obtained as follows:
\begin{align}
    \mathbf{Y}^{t}_{i} &= F_\theta(\mathbf{I}^{t}_{p}) \odot \mathbf{B}^{t}_{i}, \ \ \mathbf{Y}^{t}_{i} \in \mathbb{R}^{L_t \times C}, i \in \{1,2,\cdots,M\},
\end{align}
where $\mathbf{B}^{t}_{i} \in \mathbb{R}^{N \times 1}$ is a binary mask indicating the selected patches and $L_t$ is the number of target patches.

\textbf{Condition and Prediction.}
The three-dimensional spatial relationship between the source image and the target image is denoted as $\mathbf{a} \in \mathbb{R}^{6}$, which encapsulates the translation and rotation changes in the x, y, and z directions.
This vector $ \mathbf{a} $ is encoded by $ P_\theta $ to obtain the pose embedding, denoted as $ \mathbf{P} \in \mathbb{R}^{1 \times C} $.
Then, the 2D positional embeddings indicating the location of the context and target block are denoted as $\mathbf{Q}^{s} \in \mathbb{R}^{L_s \times C}$ and $\mathbf{Q}^{t}_i \in \mathbb{R}^{L_t \times C}, i \in \{1,2,\cdots,M\} $.
Finally, the predicted features from the world model $W_\theta$ are obtained as follows:
\begin{align}
    \mathbf{Z}^{s'} &= \mathbf{Z}^s + \mathbf{Q}^{s}, \\
    \mathbf{Z}^{t'}_i &= \mathbf{Z}^t_i + \mathbf{Q}^{t}_i, \\
    \hat{\mathbf{Y}}^{t}_{i} &= W_{\theta}(\text{concat}[\mathbf{Z}^{s'}, \mathbf{P}, \mathbf{Z}^{t'}_i]),
\end{align}
where $\mathbf{Z}^t_i \in \mathbb{R}^{L_t \times C}, i \in \{1,2,\cdots,M\} $ is a set of learnable parameters representing the features to be predicted.
These parameters interact with the context feature and condition information to generate the target feature.
Then, the models are optimized according to the loss function defined in Eq.~\eqref{eq:loss}.
\section{Experiments}

\begin{figure}[!tp]%
\centering
\includegraphics[width=\textwidth]{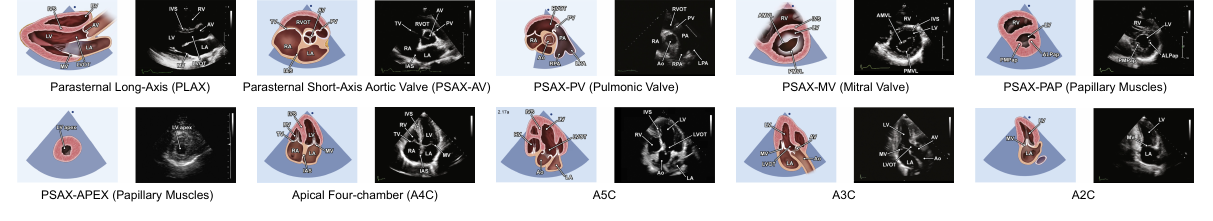}
\caption{
\textbf{Anatomic and 2D ultrasound images of ten standard planes.}
The cardiac images used in the diagram are sourced from \cite{mitchell2019guidelines}.
}
\label{fig3}
\end{figure}

\subsection{Implementation Details}
\textbf{Dataset.} In this paper, we collected data and conducted experiments on the ten most common standard planes~\cite{mitchell2019guidelines}, as shown in Fig.~\ref{fig3}.
The ultrasound images and scan data were acquired following the procedure described in \cite{jiang2024cardiac}.
Ultimately, we amassed data from 364 routine clinical scans, resulting in a total of 1.36 million image and 3D pose data pairs.
The whole data collection process was approved and supervised by The University Science and Technology Ethics Committee.
We split the dataset into 290 scans (1.07 million samples) for training and 74 scans (0.29 million samples) for testing.
For pre-training, only the training set was utilized.
Both the training and test sets were employed for the downstream probe guidance task.
It is important to note that the individuals in the training and test sets are different to avoid information leakage and fairly validate the model's generalization performance.

\textbf{Pre-training.} 
The context and target encoders were implemented using ViT-Small/16, and the world model utilized a custom vision transformer with a depth of 6 layers and a hidden dimension of 384.
The entire model was trained for 50 epochs with a batch size of 1024 on 8 Nvidia RTX-4090 GPUs. 
The training included a 7-epoch warmup period (starting learning rate is 1e-4), followed by a learning rate of 5e-4, using a cosine scheduler with a final learning rate of 5e-7.
The implementation details for generating the context block and target blocks, as well as the hyper-parameters, followed the procedures described in \cite{assran2023self}.

\textbf{Downstream task.} 
For the probe guidance task, we adopted the framework and procedure proposed in \cite{jiang2024cardiac}. 
The input for this task is an ultrasound image, and the output is the probe position adjustment needed to achieve a specific standard view.
This task aims to assist junior ultrasound medical personnel in scanning, enhancing the success rate and quality of view acquisition.
During fine-tuning, our pre-trained world model $W_\theta$ was loaded and optimized for 5 epochs with a batch size of 1024 on 8 Nvidia RTX-4090 GPUs.
The learning rate was set to 1e-4, using a cosine scheduler with a final learning rate of 1e-6.
The optimizer was set to AdamW.

\textbf{Evaluation Metrics.}
The metric used for probe guidance task is the Mean Absolute Error (MAE) between the predicted and ground truth probe poses.

\subsection{Results}
\begin{table}[!t]\small
\caption{
\textbf{Evaluation of the probe guidance task.} 
We report MAE results that represent probe guidance errors (lower is better) across ten standard planes.
}
\label{tab:probe}
\begin{center}
\resizebox{1\columnwidth}{!}{%
\begin{tabular}{p{2cm} p{3cm} p{2.2cm} p{2.2cm} p{2.2cm} p{2.2cm} p{2.2cm} p{2.4cm}}
\toprule
\multirow{2}{*}{\textbf{Plane}} & \multirow{2}{*}{\textbf{Model}} & \multicolumn{3}{c}{\textbf{Translation (mm)}} & \multicolumn{3}{c}{\textbf{Rotation (degree)}}\\
\cline{3-8}
 & & \textbf{x} & \textbf{y} & \textbf{z} & \textbf{rx} & \textbf{ry} & \textbf{rz} \\
\midrule
\multirow{2}{*}{PLAX} & Cardiac Dreamer~\cite{jiang2024cardiac} & 8.66 & 8.14  &	5.63  &	6.60  &	5.42 	 &8.23
 \\
 & \textbf{$+$ Our Pre-train} & \textbf{8.39} \textbf{\textcolor{mygreen}{(-3.15\%)}}  & \textbf{8.02} \textbf{\textcolor{mygreen}{(-1.53\%)}} & \textbf{5.53} \textbf{\textcolor{mygreen}{(-1.80\%)}} & \textbf{6.46} \textbf{\textcolor{mygreen}{(-2.12\%)}} & \textbf{5.34} \textbf{\textcolor{mygreen}{(-1.42\%)}} & \textbf{7.89} \textbf{\textcolor{mygreen}{(-4.07\%)}} \\
\midrule
\multirow{2}{*}{PSAX-AV} & Cardiac Dreamer & 7.26&	6.63&	4.51&	5.28	&6.26&	7.43 \\
 & \textbf{$+$ Our Pre-train} & \textbf{7.06} \textbf{\textcolor{mygreen}{(-2.73\%)}} & \textbf{6.57} \textbf{\textcolor{mygreen}{(-0.97\%)}} & \textbf{4.34} \textbf{\textcolor{mygreen}{(-3.85\%)}} & \textbf{5.28} \textbf{\textcolor{mygreen}{(-0.11\%)}} & \textbf{6.03} \textbf{\textcolor{mygreen}{(-3.63\%)}} & \textbf{7.18} \textbf{\textcolor{mygreen}{(-3.25\%)}} \\
\midrule
\multirow{2}{*}{PSAX-PV} & Cardiac Dreamer & 7.59	& 6.58	& 4.71	& 5.47	& 5.67	& 8.54 \\
 & \textbf{$+$ Our Pre-train} & \textbf{7.49} \textbf{\textcolor{mygreen}{(-1.38\%)}} & \textbf{6.56} \textbf{\textcolor{mygreen}{(-0.28\%)}} & \textbf{4.66} \textbf{\textcolor{mygreen}{(-1.08\%)}} & \textbf{5.46} \textbf{\textcolor{mygreen}{(-0.21\%)}} & \textbf{5.52} \textbf{\textcolor{mygreen}{(-2.73\%)}} & \textbf{8.21} \textbf{\textcolor{mygreen}{(-3.88\%)}} \\
\midrule
\multirow{2}{*}{PSAX-MV} & Cardiac Dreamer & 7.81	&6.42	&4.89	&6.68	&5.87	&9.11 \\
 & \textbf{$+$ Our Pre-train} & \textbf{7.51} \textbf{\textcolor{mygreen}{(-3.89\%)}} & \textbf{6.42} (0.04 \%) & \textbf{4.77} \textbf{\textcolor{mygreen}{(-2.53\%)}} & \textbf{6.55} \textbf{\textcolor{mygreen}{(-1.92\%)}} & \textbf{5.78} \textbf{\textcolor{mygreen}{(-1.63\%)}} & \textbf{8.78} \textbf{\textcolor{mygreen}{(-3.67\%)}} \\
\midrule
\multirow{2}{*}{PSAX-PAP} & Cardiac Dreamer & 7.18&	6.07&	4.43	&6.35	&5.39	&8.53 \\
 & \textbf{$+$ Our Pre-train} & \textbf{6.92} \textbf{\textcolor{mygreen}{(--3.57\%)}} & \textbf{5.97} \textbf{\textcolor{mygreen}{(-1.63\%)}} & \textbf{4.35} \textbf{\textcolor{mygreen}{(-1.78\%)}} & \textbf{6.11} \textbf{\textcolor{mygreen}{(-3.88\%)}} & \textbf{5.33} \textbf{\textcolor{mygreen}{(-1.04\%)}} & \textbf{8.42} \textbf{\textcolor{mygreen}{(-1.31\%)}} \\
\midrule
\multirow{2}{*}{PSAX-APEX} & Cardiac Dreamer & 6.98	&5.98	&4.25	&5.69&	4.89	&7.33 \\
 & \textbf{$+$ Our Pre-train} & \textbf{6.85} \textbf{\textcolor{mygreen}{(-1.88\%)}} & \textbf{5.77} \textbf{\textcolor{mygreen}{(-3.41\%)}} & \textbf{4.11} \textbf{\textcolor{mygreen}{(-3.28\%)}} & \textbf{5.55} \textbf{\textcolor{mygreen}{(-2.43\%)}} & \textbf{4.90} (0.07\%) & \textbf{7.25} \textbf{\textcolor{mygreen}{(-1.11\%)}} \\
\midrule
\multirow{2}{*}{A4C} & Cardiac Dreamer & 7.72& 	7.17& 	5.45& 	5.64& 	4.89& 	8.91 \\
 & \textbf{$+$ Our Pre-train} & \textbf{7.55} \textbf{\textcolor{mygreen}{(-2.14\%)}} & \textbf{7.00} \textbf{\textcolor{mygreen}{(-2.49\%)}} & \textbf{5.36} \textbf{\textcolor{mygreen}{(-1.70\%)}} & \textbf{5.48} \textbf{\textcolor{mygreen}{(-2.84\%)}} & \textbf{4.86} \textbf{\textcolor{mygreen}{(-0.55\%)}} & \textbf{8.60} \textbf{\textcolor{mygreen}{(-3.43\%)}} \\
\midrule
\multirow{2}{*}{A5C} & Cardiac Dreamer & 7.38 &6.65	&5.54	&5.83	&5.91	&12.03 \\
 & \textbf{$+$ Our Pre-train} & \textbf{7.15} \textbf{\textcolor{mygreen}{(-3.19\%)}} & \textbf{6.46} \textbf{\textcolor{mygreen}{(-2.85\%)}} & \textbf{5.40} \textbf{\textcolor{mygreen}{(-2.41\%)}} & \textbf{5.80} \textbf{\textcolor{mygreen}{(-0.49\%)}} & \textbf{5.88} \textbf{\textcolor{mygreen}{(-0.54\%)}} & \textbf{11.88} \textbf{\textcolor{mygreen}{(-1.18\%)}} \\
\midrule
\multirow{2}{*}{A3C} & Cardiac Dreamer & 7.21 &	6.52 	&5.21& 	5.92& 	6.29& 	9.81  \\
 & \textbf{$+$ Our Pre-train} & \textbf{6.90} \textbf{\textcolor{mygreen}{(-4.34\%)}} & \textbf{6.34} \textbf{\textcolor{mygreen}{(-2.81\%)}} & \textbf{5.08} \textbf{\textcolor{mygreen}{(-2.44\%)}} & \textbf{5.77} \textbf{\textcolor{mygreen}{(-2.63\%)}} & \textbf{6.12} \textbf{\textcolor{mygreen}{(-2.74\%)}} & \textbf{9.52} \textbf{\textcolor{mygreen}{(-2.98\%)}} \\
\midrule
\multirow{2}{*}{A2C} & Cardiac Dreamer & 7.28&	6.89	&4.95	&8.47	&5.46&	14.51 \\
 & \textbf{$+$ Our Pre-train} & \textbf{7.04} \textbf{\textcolor{mygreen}{(-3.32\%)}} & \textbf{6.70} \textbf{\textcolor{mygreen}{(-2.84\%)}} & \textbf{4.83} \textbf{\textcolor{mygreen}{(-2.35\%)}} & \textbf{8.37} \textbf{\textcolor{mygreen}{(-1.20\%)}} & \textbf{5.33} \textbf{\textcolor{mygreen}{(-2.45\%)}} & \textbf{13.98} \textbf{\textcolor{mygreen}{(-3.66\%)}} \\
\bottomrule
\end{tabular}
}
\end{center}
\end{table}

\begin{figure}[!tp]%
\centering
\includegraphics[width=\textwidth]{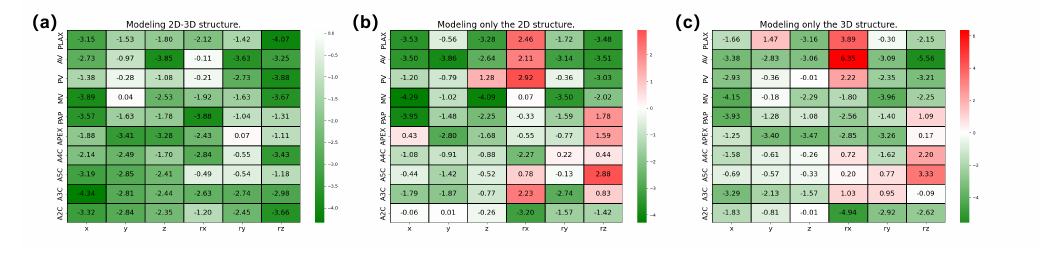}
\caption{
\textbf{Ablation of the pre-training objectives.}
The figure shows the relative change in MAE across six degrees of freedom for ten standard views, comparing different pre-training objectives with Cardiac Dreamer~\cite{jiang2024cardiac}. 
Smaller values indicate better performance.
\textbf{(a)} Our proposed 2D-3D Joint Structure-aware pre-training. 
\textbf{(b, c)} Pre-training focused only on 2D or 3D structures.
}
\label{fig5}
\end{figure}

\textbf{Comparison with SOTA.} 
To validate that the proposed structure-aware pre-training method benefits the acquisition of echocardiogram, we conducted comprehensive evaluations on the downstream probe guidance task.
As shown in Tab.\ref{tab:probe}, the pre-trained model consistently achieved better or comparable results across all dimensions in the ten standard views.
Notably, the highest observed improvement observed was up to 4.34\%.
The pre-trained model demonstrated slight weakness compared to the Cardiac Dreamer in only one dimension for the PSAX-MV and PSAX-APEX planes.
Despite this minor shortfall, the overall performance indicates that the model has effectively learned valuable information about the 2D and 3D structures of the heart.
This acquired knowledge enhances the precision of probe guidance during cardiac ultrasound scanning tasks, thereby potentially supporting less experienced sonographers in acquiring high-quality echocardiograms in the future.

\textbf{Ablations.}
To demonstrate the importance of each component in our proposed joint 2D-3D modeling approach, we decoupled the 2D and 3D modeling, either focusing solely on 2D structure or 3D structure modeling.
As shown in Fig.\ref{fig5}, our method achieves improvements in almost all dimensions, whereas using only 2D or 3D modeling results in poorer performance in rotation dimension.
In summary, while either 2D or 3D modeling alone enhances the model's performance to some extent, combining both in joint pre-training achieves the best results.
This conclusion aligns with practical experience as well. 
First, 3D modeling helps in understanding the 3D structure of the heart, while 2D modeling enables the model to more accurately identify anatomical landmarks on standard views,  thereby providing crucial guidance for probe positioning adjustments.

\section{Conclusion and Discussion}
In this work, we propose a 2D-3D joint structure-aware pre-training framework to enhance the cardiac world model's understanding of spatial relationships within two-dimensional structures on a single view and the three-dimensional spatial relationships between different views.
We innovatively designed a self-supervised learning task that predicts visual features based on both two-dimensional and three-dimensional spatial information.
To support large-scale self-supervised pre-training, we collected over a million ultrasound image and 3D pose data pairs.
After pre-training on the large-scale dataset, considerable improvement were observed in downstream probe guidance tasks across the ten standard views.
In the future, we will attempt to: 
(1) validate our pre-trained model in more downstream tasks that require a comprehensive understanding of the heart's 2D and 3D structures;
(2) validate our pre-trained model in real clinical settings, aiming to directly translate algorithmic improvements into enhanced medical outcomes or increased efficiency.

\subsubsection{Acknowledgement.}
This work was supported in part by the National Key R\&D Program of China (2021ZD0140407), the NSFC (62321005) and the Deng Feng Fund.

\subsubsection{Disclosure of Interests.}
The authors have no competing interests to declare that are relevant to the content of this article.

\bibliographystyle{splncs04}
\bibliography{mybibliography}

\end{document}